# ATTENTION TO DETAILS, LOGITS TO TRUTH: VISUAL-AWARE ATTENTION AND LOGITS ENHANCEMENT TO MITIGATE HALLUCINATIONS IN LVLMS

*Jingyi Wang, Fei Li, Rujie Liu*

Fujitsu Research and Development Center, Beijing, China
wangjingyi@fujitsu.com

## ABSTRACT

Existing Large Vision-Language Models (LVLMs) exhibit insufficient visual attention, leading to hallucinations. To alleviate this problem, some previous studies adjust and amplify visual attention. These methods present a limitation that boosting attention for all visual tokens inevitably increases attention to task-irrelevant tokens. To tackle this challenge, we propose a training-free attentional intervention algorithm to enhance the attention of task-relevant tokens based on the argument that task-relevant tokens generally demonstrate high visual-textual similarities. Specifically, the vision-text cross-attention submatrices, which represent visual-textual correlations, are extracted to construct the reweighting matrices to reallocate attention. Besides, to enhance the contribution of visual tokens, we inject visual attention values into the beam search decoding to identify solutions with higher visual attention. Extensive experiments demonstrate that this method significantly reduces hallucinations across mainstream LVLMs, while preserving the accuracy and coherence of generated content.

## 1. INTRODUCTION

Large Vision-Language Models (LVLMs) have demonstrated impressive capabilities in visual understanding tasks. Despite the content generated being coherent, these models often suffer from hallucinations, i.e., the tendency to produce nonexistent objects or incorrect details.

To mitigate hallucinations, various methods have been proposed. For example, robust instruction fine-tuning [1]-[3], and Retrieval-Augmented Generation (RAG) [4]-[5], typically incur substantial additional costs. Contrastive decoding (CD) [6]-[9] requires separate processing of the original and contrastive inputs, leading to increased inference time [10]. Another research branch focuses on attentional intervention to mitigate the inherent limitations of LVLMs [11]-[14]. These studies demonstrate two key findings: 1) the current attention mechanism does not allocate sufficient attention to visual tokens [10], [12]; 2) As more new tokens are generated, LVLMs increasingly allocate attention to text tokens [11], [15], [16]. Based on these insights, Paying Attention to Image (PAI) first proposes to enhance the attention weights for visual tokens during inference [12]. Subsequent studies have introduced improvements from various perspectives, such as selecting enhancement layers and regions [11], [16], enhancing visual attention while suppressing attention to system prompts [10].

However, they have overlooked a crucial issue. For visual understanding tasks, it requires focused attention on task-relevant image regions rather than the whole image.

Accordingly, we hope to propose an improved attentional intervention method. In visual understanding tasks, the task-irrelevant redundant information distracts models and may produce hallucinations [17]. We empirically notice that the task-relevant tokens typically exhibit higher visual-textual similarities in visual understanding tasks. Based on this observation, we introduce a novel plug-and-play method referred to as Visual-Aware Attention and Logits Enhancement (VAALE). VAALE consists of two modules, attention refocusing and visual beam search. The attention refocusing module refocuses attention on task-relevant tokens, i.e., the tokens with high visual-textual similarities. We are inspired by the technology of self-augmentation via self-reweighting (SASR) for language models, which aims to strengthen attention for semantically similar tokens across sentence pairs [18]. We inherit the pipeline of construction of reweighting matrices which function as attention augmentation signals. Besides, several modifications are made to the method to make it applicable to LVLMs, which will be discussed in the following section. Visual beam search strategy enhances the contributions of visual tokens to the outputs during inference. It selects the most vision-dominant candidate as the final outputs by adding the original logit scores with the weighted visual attention values. These two modules are both plug-and-play and can be utilized independently.

Experimental results on LLaVA-v1.5 and Qwen2.5-VL validate the capabilities of hallucination mitigation of VAALE. Compared to the second-best baseline, our method reduces $CHAIR_I$ and $CHAIR_S$ by 15.54% and 17.80%, respectively, without omitting essential details.

In summary, the main contributions are as follows:
- We introduce an attention refocusing module to refocus attention on task-relevant tokens. First, identifying task-specific tokens is transformed into highlighting tokens with high visual-textual similarities. Then it extracts the cross-attention submatrices, which represent the interaction between the visual and textual tokens, to dynamically reallocate attention.
- Moreover, we propose a visual beam search module to boost the role of visual tokens. The weighted visual attention values are incorporated into the original logit scores, thereby mitigating the imbalance of attention among visual and textual tokens.

Our method is training-free and plug-and-play. Experiments on standard hallucination benchmarks demonstrate its effectiveness in mitigating hallucinations.

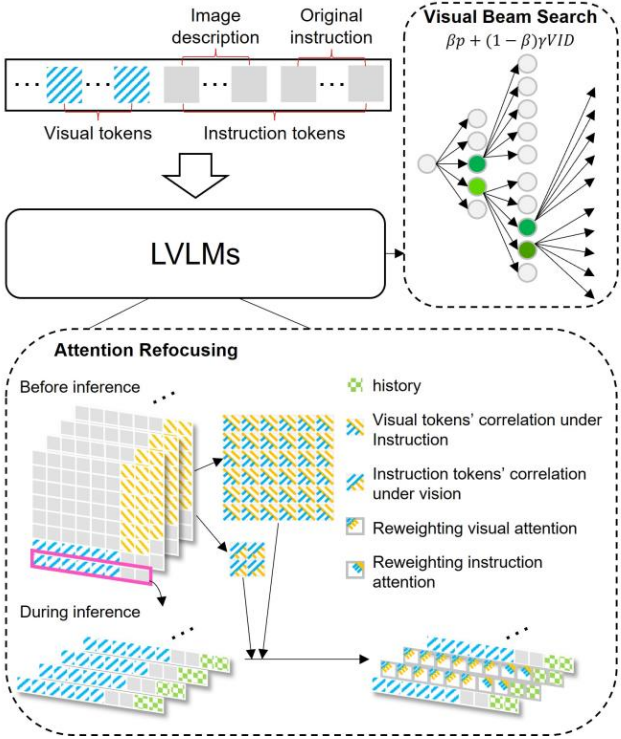

Fig. 1. Architecture of VAALE. Attention Refocusing enhances the attention of task-relevant tokens. Visual Beam Search strengthens the role of image during decoding.

## 2. METHOD

Inspired by the empirical cognition: task-relevant tokens commonly demonstrate greater visual-textual alignment in visual understanding tasks, we introduce attention refocusing mechanism, which utilizes the extracted cross-attention submatrices (showing how visual and text tokens interact) as supplementary evidences, incorporating them with the original attention weights at the selected layers. This mechanism effectively enhances the attention of tokens exhibiting high visual-textual alignment, which essentially boosts the attention devoted to task-relevant tokens. Moreover, to mitigate the imbalance in attention between visual and textual tokens, a visual beam search decoding strategy is introduced to inject weighted visual attention into the logit scores to enhance the contribution of visual tokens. Figure 1 provides an overview of the architecture of VAALE. All details are listed below.

### 2.1. Attention Refocusing

In this section, we formulate the proposed Attention Refocusing process on LVLMs to facilitate an easier understanding.

**Notation.** We simplify the input sequences, which are organized in the order of visual tokens and instruction tokens, denoted as $[X_v, X_i]$ before the generation. For this input sequence to a specific layer, the attention matrix before softmax operation is calculated as follows:

$$A = \frac{Q \cdot K^T}{\sqrt{d}} \quad (1)$$

$$Q \cdot K^T = \begin{bmatrix} f_Q(X_v) \\ f_Q(X_i) \end{bmatrix} \cdot \begin{bmatrix} f_K(X_v)^T & f_K(X_i)^T \end{bmatrix}$$
$$= \begin{bmatrix} f_Q(X_v) \cdot f_K(X_v)^T & f_Q(X_v) \cdot f_K(X_i)^T \\ f_Q(X_i) \cdot f_K(X_v)^T & f_Q(X_i) \cdot f_K(X_i)^T \end{bmatrix} \quad (2)$$

where $f_Q$ and $f_K$ are the projection matrices for the queries and keys, and $d$ is the hidden dimensions.

SASR performs attentional intervention by two steps: self-reweighting and self-augmentation. Drawing inspiration from SASR, we explore the implementation strategies of how it can be adapted and applied to LVLMs.

We define the correlation matrices as follows:

$$W_v = f_Q(X_v) \cdot f_K(X_i)^T \cdot f_Q(X_i) \cdot f_K(X_v)^T \quad (3)$$

$$W_i = f_Q(X_i) \cdot f_K(X_v)^T \cdot f_Q(X_v) \cdot f_K(X_i)^T \quad (4)$$

where $f_Q(X_v) \cdot f_K(X_i)^T \in \mathbb{R}^{l_v \times l_i}$ and $f_Q(X_i) \cdot f_K(X_v)^T \in \mathbb{R}^{l_i \times l_v}$ represent the visual-to-instruction and the instruction-to-visual cross-attention sub-matrices extracted from the full attention matrix in Equation (2), respectively. These sub-matrices capture interactions between the instruction tokens (length $l_i$) and the visual tokens (length $l_v$). $W_v \in \mathbb{R}^{l_v \times l_v}$ and $W_i \in \mathbb{R}^{l_i \times l_i}$ compute their similarities, thus integrating visual and textual information with each other.

For visual understanding tasks, we introduce a simple strategy to ensure semantic relevance between visual tokens and instruction tokens. As shown in Figure 1, we first use a fixed instruction, i.e., "Please describe this image in detail.", to generate a description of the image. Then we concatenate the generated description before the original instruction. Thus, the new prompt is guaranteed to contain tokens semantically aligned with the visual tokens.

Based on the correlation matrices, we perform attention reweighting on the attention weights matrix. Unlike SASR, we make customized design for LVLMs. Since the model's final vocabulary probability distribution is generated by projecting the hidden states of the sequence's last token, we reallocate the attention weights of the last token. Moreover, in current LVLMs using a cache to store precomputed hidden states to reduce latency and computation cost, the attention matrix remains a full square array before inference but collapses to a one-dimensional vector during inference when only the last token is active. Accordingly, $W_v$ and $W_i$ are stored before inference and are loaded during inference. Thus, self-reweighting can be computed as:

$$R_v = W_v \cdot A_v \quad (5)$$
$$R_i = W_i \cdot A_i \quad (6)$$

where $R_v \in \mathbb{R}^{1 \times l_v}$ and $R_i \in \mathbb{R}^{1 \times l_i}$ represent the reweighting attention matrices of the visual tokens and instruction tokens, respectively. $A_v$ and $A_i$ represent the original visual attention and instruction attention, respectively.

Subsequently, the reweighting attention matrices are utilized as an augmentation signal to refocus the attention weights on the parts where there is a significant cross-modal semantic association.

Guided by [11], attention weights are adjusted within the layers where the generated token exhibits strong interactions with visual information while demonstrating weak correlations with the corresponding image patches. Formally, for each visual token and instruction token across all heads in the selected layers, the results of attention refocusing can be computed as:

$$A_v = R_v + \alpha A_v \quad (7)$$
$$A_i = R_i + \alpha A_i \quad (8)$$

where $\alpha > 0$ denotes the hyperparameter defined as a balance factor to control the ratio between the reweighting attention matrices and the original attention weights.

## 2.2. Visual Beam Search

To mitigate the hallucinations due to insufficient attention on visual information, we propose the visual beam search decoding method. In our framework, a response that interacts more with image is considered more reliable.

We define the Visual Interaction Degree (VID) between the generated last token $y_k$ and the image in each head as:

$$VID(y_k) = \frac{1}{(l_2 - l_1)} \sum_{l=l_1}^{l_2} \sum_{i=1}^{n} A_v^k(i) \quad (9)$$

where $VID(y_k) \in \mathbb{R}^{n_{beam} \times 1}$, and $n_{beam}$ represents the number of beams; $A_v^k(i)$ represents the cross-attention weights between $y_k$ and the $i$-th visual token; $l_1$ and $l_2$ represent the start layer and the end layer that are used to compute VID. Here we select the layers where visual information interaction primarily occurs according to [11].

Due to sequential coherence, adjacent tokens typically exhibit correlated attention patterns [19]. Accordingly, the logit scores of the next predicted token $y_{k+1}$ in each head can be updated by:

$$p(y_{k+1}) = \beta \cdot p(y_{k+1}) + (1-\beta) \cdot \gamma \cdot VID(y_k) \cdot 1_{L_{vocab}}^T \quad (10)$$

$$1_{L_{vocab}}^T = [1, 1, \cdots, 1] \in \mathbb{R}^{1 \times L_{vocab}} \quad (11)$$

where $p(y_{k+1}) \in \mathbb{R}^{n_{beam} \times L_{vocab}}$ represents the predicted logit scores of $y_{k+1}$, and $L_{vocab}$ represents the length of vocabulary. The weight $\beta$ and $\gamma$ are the hyperparameter used to control the degree of dependence on the visual information and the scaling coefficient of VID, respectively.

## 3. EXPERIMENTS

### 3.1. Setup and results

**Implementation Details.** We conduct experiments on two representative LVLMs, i.e., LLaVA-v1.5-7B [20] and Qwen2.5-VL-3B [21], to evaluate the effectiveness of our method. Attention Refocusing contains two parameters: the range of the layers selected to refocus attention and the balance factor $\alpha$ in Equation (7-8). Visual beam search contains three parameters: the range of the layers selected to calculate VID, the balance factor $\beta$ and the scaling coefficient $\gamma$ in Equation (10). In our experiments, we unify $N_{beam}$=5 for all the beam search decoding-based methods. The detailed settings are listed in Table 1.

Table 1. Settings of hyper-parameters

| Model | AR layers | $\alpha$ | VID layers | $\beta$ | $\gamma$ |
|---|---|---|---|---|---|
| LLaVA-v1.5-7B | [5, 18] | 0.4 | [5, 26] | 0.4 | 0.15 |
| Qwen2.5-VL-3B | [10, 30] | 0.2 | [10, 34] | 0.4 | 0.3 |

**Baselines.** We compare VAALE with three mainstream hallucination mitigation approaches as the baseline methods, including OPERA [13], VCD [6], and PAI [12]. OPERA improves beam search by introducing an over-trust penalty term on the attention weights during inference with a rollback strategy to mitigate hallucinations. VCD is a nucleus sampling-based decoding method, which contrasts output distributions from both original and perturbed visual inputs to reduce the statistical priors. PAI first adjusts the attention matrix, followed by contrastive decoding to further alleviate language bias. In our experiments, we use the default hyperparameters of these baselines.

**Benchmark and Metrics.** Following [12], we conduct our evaluation with 500 random images from the COCO 2014 validation set. We use two types of criteria, i.e., Caption Hallucination Assessment with Image Relevance (CHAIR) [22] and Polling-based Object Probing Evaluation (POPE) [23] to evaluate the degree of hallucinations.

CHAIR evaluates how well the generated captions align with the content of the given image. It includes instance-level ($CHAIR_I$) and sentence-level ($CHAIR_S$), defined as follows:

$$CHAIR_I = \frac{|\{hallucinated\ objects\}|}{|\{all\ mentioned\ objects\}|} \quad (12)$$

$$CHAIR_S = \frac{|\{captions\ w/\ hallucinated\ objects\}|}{|\{all\ captions\}|} \quad (13)$$

We also employ F1 scores to evaluate the richness and accuracy of the generated description. We adopt the same settings of the max new token to 512 as used in [11]-[13].

POPE is a binary VQA-based metric which assesses hallucinations by asking LVLMs questions (e.g. "Is [object] in the image?"). Moreover, to emphasize that this is a binary VQA task, the prompt "Please answer yes or no" is appended to the original question. The ground-truth object follows three different sampling options: random, popular and adversarial.

**Results.** Table 2 presents the experimental results of the CHAIR evaluation. From Table 2, VAALE achieves a significant improvement compared to the vanilla LVLMs with both greedy and beam search decoding methods. Compared to the second-best baseline PAI, VAALE still reduces hallucinations by 15.5% in $CHAIR_S$ and 17.8% in $CHAIR_I$ for LLaVA-1.5-7B. Besides, our proposed achieves higher F1 scores, demonstrating its effectiveness on generating rich and accurate descriptions. For Qwen2.5-VL, which has not been integrated by the baseline approaches, we reduce the $CHAIR_S$/ $CHAIR_I$ metrics from 29.4/7.7 to 17.8/4.4 while maintaining the F1-score at a comparable level.

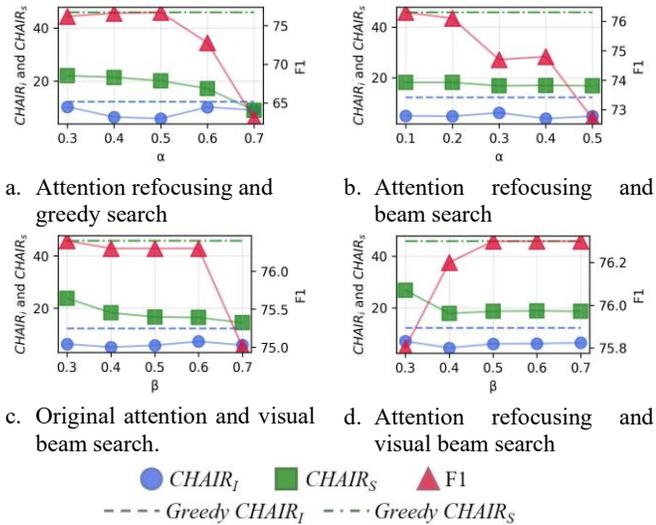

a. Attention refocusing and greedy search

b. Attention refocusing and beam search

c. Original attention and visual beam search.

d. Attention refocusing and visual beam search

Figure 2 Ablation study of hyperparameters.

Table 2 CHAIR hallucination evaluation and F1 results on two LVLMs. VAALE demonstrates significant hallucination alleviation compared to other baselines while maintaining the accuracy and completeness of the generated captions.

| Method | $CHAIR_S$ ↓ | $CHAIR_I$ ↓ | F1 ↑ |
|---|---|---|---|
| LLAVA-1.5-7B | | | |
| Greedy | 53.0 | 15.6 | 76.7 |
| Beam search | 55.6 | 15.4 | 77.5 |
| OPERA | 45.6 | 13.1 | 79.1 |
| VCD | 58.6 | 18.2 | 72.8 |
| PAI | 29.6 | 7.3 | 76.0 |
| VAALE | **25.0** | **6.0** | **78.0** |
| | (↓15.54%) | (↓17.80%) | |
| Qwen2.5-VL-3B | | | |
| Greedy | 45.8 | 12.2 | 76.1 |
| Beam search | 29.4 | 7.7 | 75.9 |
| VAALE | **17.8** | **4.4** | **76.2** |
| | (↓39.45%) | (↓42.85%) | |

Table 3 shows the performance on the POPE benchmark. Compared to the baselines, our method achieves notable improvements.

## 3.2. Ablation study

To demonstrate the efficacy across the modules we introduced in our VAALE, we conduct in-depth ablation studies as detailed in Figure 2. We use Qwen2.5-VL-3B as the LVLM baseline, and the greedy decoding and the beam search decoding methods as the decoding baselines to evaluate performance on the task of image description. The CHAIR metric is used for evaluation.

Table 3 Performance on POPE. Results are averaged across the three sampling options (i.e., random, popular and adversarial).

| Method | Acc | F1 |
|---|---|---|
| LLAVA-1.5-7B | | |
| Greedy | 84.61 | 85.30 |
| Beam search | 85.08 | 85.63 |
| OPERA | 84.21 | 83.55 |
| VCD | 82.27 | 83.16 |
| PAI | 85.82 | 85.97 |
| VAALE | **86.11** | **86.83** |
| Qwen2.5-VL-3B | | |
| Greedy | 89.09 | 88.22 |
| Beam search | 89.34 | 88.69 |
| VAALE | **89.61** | **89.36** |

**Effect of attention refocusing.** Figure 2a and Figure 2b show how CHAIR metrics and F1 score vary with the balancing factor $\alpha$ under greedy search and beam search decoding, respectively. We find that setting $\alpha$ between 0.3 and 0.5 can improve performance under greedy search, with the optimal value around 0.5 balancing hallucination reduction and content richness. Under beam search, the valid range of $\alpha$ lies between 0.1 and 0.2, with the optimal value around 0.2. The results suggest that independent attention refocusing can consistently mitigate hallucinations of LVLMs under different decoding strategies.

**Effect of visual beam search.** Figure 2c shows the performance under different coefficients (i.e., $\beta$) modulating reliance on visual cues. The scaling coefficient $\gamma$ is fixed as Table 1. From Figure 2c, we observe that performance gains are achieved by setting $\beta$ between 0.3 and 0.6, with the optimal value around 0.4. This validates the capability of the visual beam search strategy to alleviate hallucinations.

Finally, we combine attention refocusing with visual beam search. Figure 2d presents the dependence of the metrics on $\beta$ at the fixed best $\alpha$ value of 0.2. The results show that the combination of two strategies can further improve the richness of descriptions while preserving the ability to effectively mitigate hallucinations.

## 4. CONCLUSION

In this paper, we argue that visual-textual semantically similar tokens are generally task-relevant and insufficient visual attention is one reason for hallucinations of LVLMs. Based on these findings, we propose VAALE, a training-free method that mitigates hallucinations of LVLMs by refocusing attention on task-related tokens (i.e., visual-textual semantically similar tokens) and injecting visual information into beam search decoding. Extensive experiments across multiple benchmarks and LVLMs demonstrate the effectiveness of VAALE in alleviating hallucinations.